\documentclass[letterpaper]{article} 
\usepackage{aaai2026}  
\usepackage{times}  
\usepackage{helvet}  
\usepackage{courier}  
\usepackage[hyphens]{url}  
\usepackage{graphicx} 
\usepackage{natbib}  
\usepackage{caption} 
\usepackage{algorithm}

\usepackage{newfloat}
\usepackage{listings}
\DeclareCaptionStyle{ruled}{labelfont=normalfont,labelsep=colon,strut=off} 
\floatstyle{ruled}
\newfloat{listing}{tb}{lst}{}
\floatname{listing}{Listing}
\usepackage{booktabs}
\usepackage{algpseudocode}
\usepackage{algpseudocode}
\usepackage{enumitem} 
\usepackage{multirow}

\title{DanceHA: A Multi-Agent Framework for Document-Level Aspect-Based Sentiment Analysis}
\author{
    Lei Wang\textsuperscript{\rm 1}\thanks{Corresponding author.},
    Min Huang\textsuperscript{\rm 2},
    Eduard Dragut\textsuperscript{\rm 1}
}

\affiliations{
    \textsuperscript{\rm 1}Temple University, Philadelphia, PA, USA \\
    \textsuperscript{\rm 2}Independent Researcher, Philadelphia, PA, USA \\
    \{tom.lei.wang, edragut\}@temple.edu, minerstudy@gmail.com
}

\usepackage{bibentry}

\begin{document}

\maketitle

\begin{abstract}
Aspect-Based Sentiment Intensity Analysis (ABSIA) has garnered increasing attention, though research largely focuses on domain-specific, sentence-level settings. In contrast, document-level ABSIA--particularly in addressing complex tasks like extracting Aspect-Category-Opinion-Sentiment-Intensity (ACOSI) tuples--remains underexplored. In this work, we introduce DanceHA, a multi-agent framework designed for open-ended, document-level ABSIA with informal writing styles. DanceHA has two main components: Dance, which employs a divide-and-conquer strategy to decompose the long-context ABSIA task into smaller, manageable sub-tasks for collaboration among specialized agents; and HA, Human-AI collaboration for annotation. We release Inf-ABSIA, a multi-domain document-level ABSIA dataset featuring fine-grained and high-accuracy labels from DanceHA. Extensive experiments demonstrate the effectiveness of our agentic framework and show that the multi-agent knowledge in DanceHA can be effectively transferred into student models. Our results highlight the importance of the overlooked informal styles in ABSIA, as they often intensify opinions tied to specific aspects. 
\end{abstract}

\begin{links}
\link{Code}{https://github.com/Tom-Owl/DanceHA}
\end{links}

\section{Introduction}
Aspect-Based Sentiment Analysis (\textbf{ABSA}) is a fine-grained sentiment analysis \cite{schneider2018debugsl, aljebreen2021segmentation} direction that aims to identify sentiments expressed toward specific \textit{aspects} (or attributes) of an entity within a given text  \cite{SemEval14-Pontiki-ACL14, SemEval16-Pontiki-ACL16}. It refers to the process of identifying the aspect terms (or categories) mentioned (e.g., ``battery life'') in a product review and determining the \textit{sentiment polarity} (e.g., positive, negative, neutral) associated with each aspect \cite{DragutYW10}.  
One task in ABSA is Aspect-Category-Opinion-Sentiment (\textbf{ACOS}) tuples extraction \cite{ACOS-Cai-21ACL}, where in addition to aspect and sentiment, it seeks to extract broader category class to which the aspect belongs (e.g., hardware for battery) and the subjective expression \textit{opinion} \cite{schneider2015towards, hosseinia2019pro}, such as amaaaazing and dull. Consider the sentence ``The battery life of this laptop is amaaaazing, but the screen is a bit dull.'' There are 2 ACOS tuples, (battery life, hardware, amaaaazing, positive) and (screen, display, a bit dull, negative). The prior work of ABSA span both small language models (SMLs) and large language models (LLMs) \cite{BertABSA-Hoang-19ACL, Aspect-zhang-21EMNLP, LEGO-gao-22ICCL, MvpABSA-Gou-ACL23, RefineABSA-Su-24ACL, llama-Šmíd-24WASSA, CompoundABSALLM-Bai-24EMNLP}. 

Recent research on ABSA aims to
also quantify the intensity or strength of those sentiments (e.g., mild, strong, extreme). Like sentiment, the representation of intensity can be represented ordinal, categorical, or continuous. In this study, we use a 5-point Likert scale for intensity, where 0 represents neutral and 5 represents extreme. In our example, the intensities are 5 for “amaaaazing” and 2 for “a bit dull,” respectively.
This is know as Aspect-Based Sentiment Intensity Analysis (\textbf{ABSIA}) \cite{Service-Mamta-23ECIR}. ACOS is thus extended to include intensity, leading to the extraction of the tuple Aspect-Category-Opinion-Sentiment-Intensity (\textbf{ACOSI}).
Both ABSA and ABSIA have been studied at sentence level. Document-level ABSA and ABSIA are newer, more challenging tasks, and relatively underexplored \cite{Sequence2Structure-Song-23EMNLP, DocumentABSA-Kasturi-23arXiv}, primarily due to the lack of datasets \cite{ChallengesTSA-Luo-22Arxiv}. 
A key challenge in those problems
is the difficulty of constructing large-scale, high-quality, fine-grained labeled datasets, which are costly and labor-intensive to produce.
\textit{In this work, we investigate whether multi-agent frameworks can effectively tackle document-level ACOSI.}

The main focus of ABSA is to analyze user-generated content (UGC), such as product reviews \cite{yang2017satirical, he2020dynamics, ACOS-Cai-21ACL, chen2025comcrawler}.
However, there is a distinctive but overlooked feature of UGCs, which is the informal linguistic styles, such as emoji \cite{IncorporatEmoji-Singh-19ACL}
and lengthening words, e.g., 'coool' and 'goooood!!!!'. \citet{Brody2011cool} define lengthening words as the addition of extra characters to a word’s standard spelling to emphasize or alter its meaning.
While previous work highlights lengthening words as important expressions in sentiment analysis \cite{overlooked-wang-24EMNLP}, none studies them in the context of ABSA and ABSIA. 
\textit{We investigate the contributing role of lengthening words 
in Document-level ABSIA in this paper.}

\begin{figure*}[t]
\centering
\includegraphics[width=0.85\textwidth]{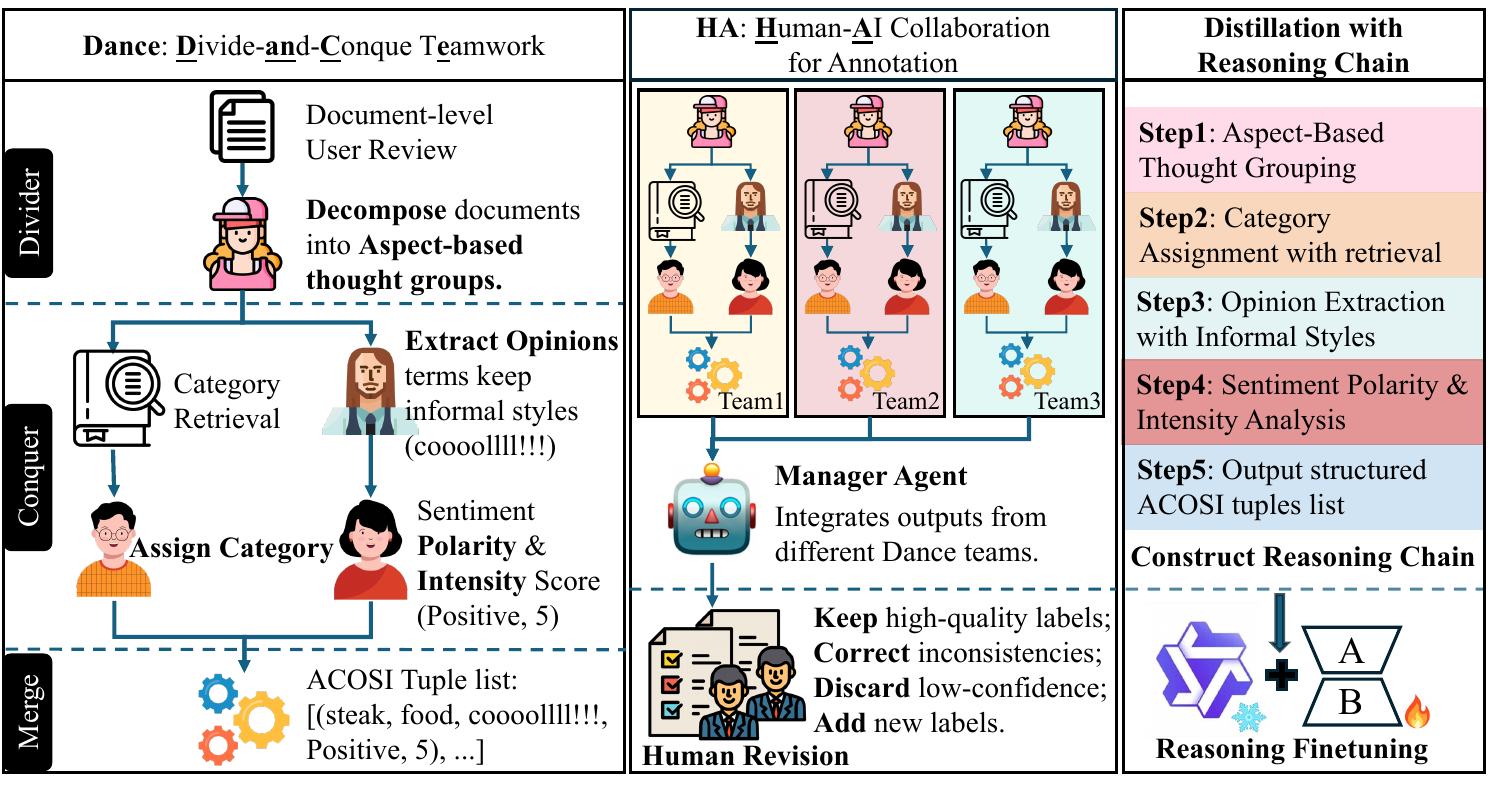}
\caption{Overview of \textbf{DanceHA}, consisting of two key components: (1) \textbf{Dance} (\underline{D}ivide-\underline{an}d-\underline{C}onquer T\underline{e}amwork) for open-ended document-level ABSIA and (2) \textbf{HA} (\underline{H}uman-\underline{A}I collaboration) for label annotation. Figure \ref{fig:Dance_case} illustrates  Dance with an example. 
}
\label{fig:AgentFramework}
\end{figure*}

In this work, we aim to propose a multi-agent framework designed for open-ended document-level ABSIA with informal styles as shown in Figure \ref{fig:AgentFramework}. 
Specifically, our agentic framework, \textbf{DanceHA}, consists of two key components: \textbf{Dance} (\underline{D}ivide-\underline{an}d-\underline{C}onquer T\underline{e}amwork), which decomposes the complex document-level ABSIA task into aspect-based thought groups for collaboration of expert agents, and \textbf{HA} (\underline{H}uman-\underline{A}I collaboration for annotation), which is a pipeline where two Manager Agents integrate outputs from multiple Dance teams, while human annotators further evaluate and revise the labels automatically generated by the Manager Agents.
We curated the \textbf{Inf-ABSIA} (\underline{Inf}ormal \underline{ABSIA}) dataset, originally an unlabeled document-level dataset from 3 publicly available datasets. 
We use DanceHA to annotate the unlabeled documents with fine-grained labels. 
Human evaluation confirms the quality of the annotations.
We further explore knowledge distillation with reasoning chains constructed with Inf-ABSIA.
Extensive experiments on 3 domains and 7 LLMs show that our proposed agentic framework achieves strong performance on long-context, open-ended ABSIA tasks. 
Our contributions are as follows:
\begin{itemize}\setlength\itemsep{0em}
\item We present \textbf{DanceHA}, a multi-agent framework with human-in-the-loop \cite{zhang2019invest, zhang2023human} designed for document-level ABSIA with informal and open-ended settings. Our empirical results show that Dance significantly outperforms few-shot CoT, and the Manager agent can further achieve performance boost.

\item We create \textbf{Inf-ABSIA}, a novel ABSIA dataset consisting of 2,714 long-context documents across three domains. Each document contains an average of 8.48 high-quality ACOSI tuples, offering a valuable resource for studying ABSIA at the document level and enabling future research on long-form opinion mining.

\item We show the importance of informal styles in ABSIA, showing that informal expressions convey stronger sentiment intensity toward specific aspects and opinions. 

\item We show distillation with reasoning chains can effectively transfer knowledge in DanceHA into student models. Our reasoning fine-tuned Qwen-14B outperforms few-shot CoT GPT-4o across all domains and achieves performance comparable to Dance with Qwen2.5-72B.
\end{itemize}

\section{Related Work}
\noindent \textbf{LLM-based Multi-Agent Collaboration}
LLM-based agentic framework \cite{zhu2025overhearing, feng2025rethinkingllmuncertaintymultiagent} have gained attention due to their ability to simulate human-like reasoning, collaboration, and decision-making \cite{LanguageAgents-Su-24EMNLP, zhu2023calypso}. A line of research in this area is multi-agent collaboration for domain-specific tasks such as Question Answering (QA) \cite{zhao2025sciarena, yifei2025researchqa}, Peer Review \cite{AgentReview-Jin-24EMNLP}, and script evaluation and generation \cite{ABSEval-Liang-24EMNLP, dugan-etal-2024-raid}.
To improve multi-agent teamwork, some studies employ debate strategies \cite{Encouraging-Liang-24EMNLP}, while others adopt decomposition strategies to guide agents in solving complex tasks \cite{DARA-Fang-24EMNLP, ChainofAgents-Zhang-24NeurIPS}. 
Building on these methods, we propose DanceHA, a training-free multi-agent framework for document-level
ACOSI tuple extraction.

\noindent \textbf{LLM-based ABSA}
Sentence-level ABSA is an old task,
yet it continues to pose challenges to LLMs \cite{MvpABSA-Gou-ACL23, CompoundABSALLM-Bai-24EMNLP}. 
\citet{llama-Šmíd-24WASSA} proposes instruction-tuning Orca-2 on domain-specific datasets.
Recent studies also begin to address more complex, document-level ABSA subtasks \cite{Sequence2Structure-Song-23EMNLP}. However, progress in this area remains limited due to the lack of fine-grained labeled datasets for tasks such as ACOS tuple extraction.
To mitigate this, \citet{CompoundSplitABSA-Seo-24EMNLP} propose a sentence-splitting strategy that segments compound sentences by aspect-term boundaries, demonstrating gains on sentence-level benchmarks.
In contrast, our work targets the open-ended document-level ACOSI tuple extraction task. We introduce a multi-agent collaboration framework that integrates sentiment intensity analysis to generate fine-grained annotations in the absence of manually labeled data.
\section{Dataset Construction}
\label{sec:InformalABSIA_dataset}
We introduce \textbf{Inf-ABSIA} (\underline{Inf}ormal \underline{ABSIA}), a document-level dataset  for ABSIA with informal styles.
Inf-ABSIA is built from 3 publicly available datasets across 3 domains:
\textbf{Laptop} (Lap) from Amazon Reviews-Electronics \cite{ni2019justifying} 
which contains user reviews for different electronic products.  We use LLM-as-Judge \cite{JudgingLLMs-Zheng-23NeurIPS} to decide whether a review belongs to laptop domain. 
\textbf{Restaurant} (Rest) from Yelp \cite{yelpdataset} with reviews under restaurants topic \cite{Xu-BertPost-19ACL}, and \textbf{Hotel} from TripAdvisor \cite{li2020hotel}.

For each domain, we randomly sample 910 review documents that contain lengthening words, identified using a regex-based tool from prior work \cite{zhang2018regular, overlooked-wang-24EMNLP}. 
Table \ref{tab:cmp_absa_dataset} compares our Inf-ABSIA with existing sentence-level and document-level ABSA and ABSIA datasets across multiple dimensions. Note that the original datasets lack ABSIA labels. We annotate them with our proposed framework, \textbf{DanceHA}. Compared to existing datasets, Inf-ABSIA spans more domains and has a larger sample size (2,714 vs 1,194 documents), longer context (average 90 vs 60 words per document), and significantly more fine-grained annotations (23,024 vs. 4,464 tuples).

\begin{table*}[t]
\centering
\resizebox{0.95\textwidth}{!}{
\begin{tabular}{lclllccccccc}
\toprule
\textbf{Dataset} &  \textbf{\# D} & \textbf{Level} & \textbf{Topic} & \textbf{Lang} & \textbf{ACOS} & \textbf{Intensity} & \textbf{Informal} & \textbf{Size} & \textbf{Avg \# w} & \textbf{Avg \# Tuple} & \textbf{\# Tuple}\\
\hline
SemEval2016-SB1 \cite{SemEval16-Pontiki-ACL16} & 2 & Sentence & ABSA & E & Y & N & N& 5,984 & 13 & 1.18 & 7,061\\ 
ACOS-Rest\&Lap \cite{ACOS-Cai-21ACL} & 2 & Sentence & ABSA & E & Y & N & N & 6,362 & 16 & 1.48 & 9,415 \\ 
ABSIA \cite{Service-Mamta-23ECIR} & 1 & Sentence & ABSIA & E & N & Y & N & 4,650 & 17 & 1.00 & 4,650\\
SIGHAN2024 \cite{Overview-lee-24sighan} & 1 & Sentence & ABSIA & C & Y & Y & N & 8,150 & 15 & 1.50 & 12,225 \\
 \midrule
SemEval2016-SB2 \cite{SemEval16-Pontiki-ACL16} & 2 & Document & ABSA & E & N & N & N & 900 & 60 & 4.96 & 4,464 \\ 
DOTSA \cite{ChallengesTSA-Luo-22Arxiv} & 6 & Document & ABSA & E & N & N & N & 1,194 & 23 & 1.26 & 1,504\\
\textbf{Inf-ABSIA (Ours)} & 3 & Document & ABSIA & E  & Y & Y & Y & \textbf{2,714} & \textbf{90} & \textbf{8.48} & \textbf{23,024} \\
\bottomrule
\end{tabular}
}
\caption{Statistics comparison of Inf-ABSIA with existing ABSA/ABSIA datasets. Inf-ABSIA has larger sample size (2,714 vs. 1,194), longer context (avg. words 90 vs. 60), and more fine-grained labels (avg. tuples 8.48 vs. 4.96). \# D indicates the number of domains. \# w denotes the number of words. Lang denotes the language, where E stands for English and C for Chinese.}
\label{tab:cmp_absa_dataset}
\end{table*}

\section{Method}
\label{sec:DanceMA}
We present an overview of our proposed framework \textbf{DanceHA} in Figure~\ref{fig:AgentFramework}.
Formally, we denote a long-context document by $d$ from a specific domain $dm \in \{Rest, Lap, Hotel\}$. The goal is to perform ACOSI tuple extraction, producing a structured list where each element follows the format \((a, c, o, s, i)\), where $a$ is an aspect term extracted from $d$, $c$ is a category from a pre-defined list, $o$ is the associated opinion word or phrase from $d$, $s$ indicates the sentiment polarity positive/negative, and $i$ is the sentiment intensity score (SIS).

\begin{figure}[t]
\centering
\includegraphics[width=0.45\textwidth]{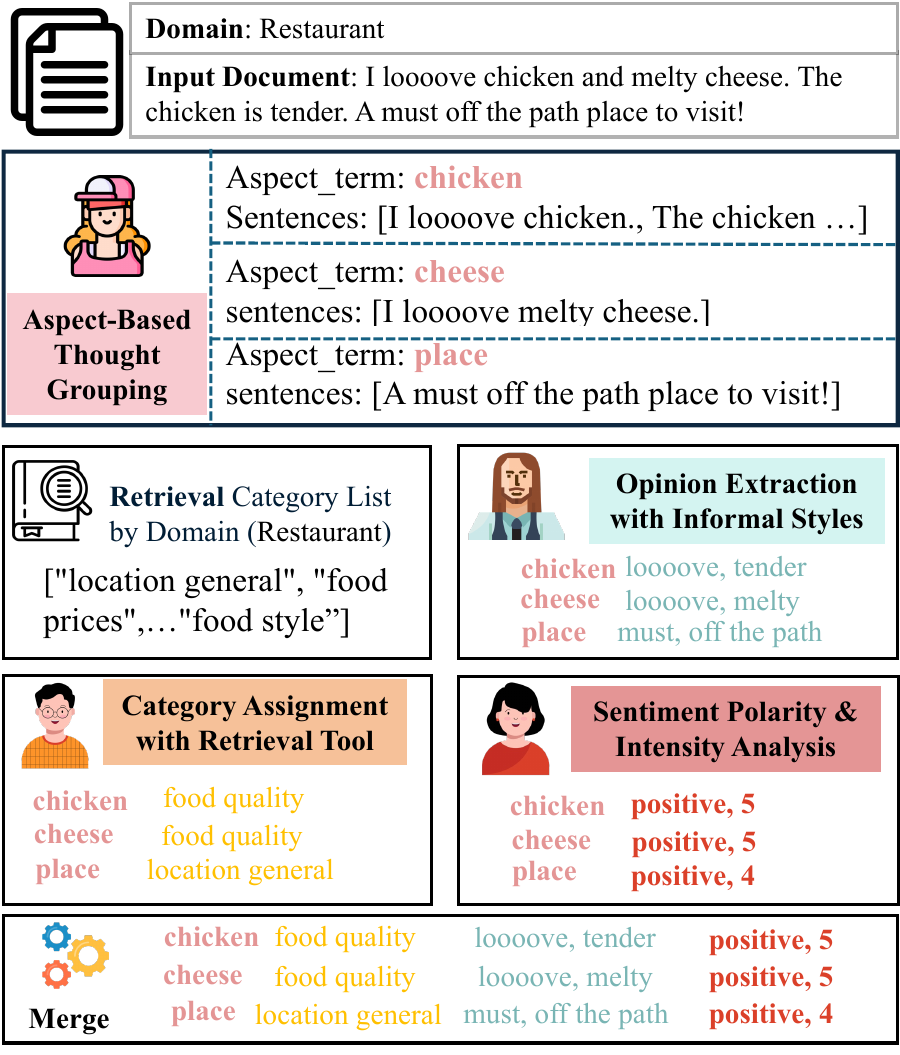}
\caption{An example of \textbf{Dance} for document-level ABSIA.
}
\label{fig:Dance_case}
\end{figure}

\subsection{Dance: Divide-and-Conquer Teamwork for Document-level ABSIA}
Dance is a multi LLM-based agents framework built on the divide-and-conquer principle, designed to handle document-level ABSIA with informal styles through agentic collaboration.
As shown in Figure \ref{fig:Dance_case}, Dance operates in three stages. In Stage 1, the Divider agent segments long documents into manageable aspect-based thought groups. Each thought group is a concise, semantically meaningful unit--typically a few sentences--representing a single aspect, facilitating  a clearer understanding of the content.
In stage 2, the thought groups are processed collaboratively by 3 specialized LLM-based agents, each responsible for one of the following tasks: Category Assignment, Opinion Extraction, and Sentiment and Intensity Analysis.
In stage 3, Dance merges the outputs from all agents with a rule-based integration method to form a structured list of ACOSI tuples.

\noindent \textbf{Divider: Aspect-Based Thought Grouping}
The Divider agent $F_D$ decomposes a 
document $d$ into structured aspect-based thought groups $\{(a_0, g_0), \ldots, (a_n, g_n)\}$, where $a$ is an aspect term and $g$ is the corresponding thought group. 
Based on the work of \cite{CompoundSplitABSA-Seo-24EMNLP}, the design of Divider agent has three steps: 1) Split: Split segment sentences containing multiple aspects into smaller thought groups, each corresponding to a distinct aspect $a$. 2) Combine: Collect consecutive sentences that pertain to the same aspect into one thought group $g$. 3) Extract Aspect: Identify and extract explicit aspect terms from the thought group. For convenience, we abbreviate the set \(\{(a_0, g_0), \ldots, (a_k, g_k)\}\) as \(\{(a, g)\}\), and this style will be used hereafter. The process can be formed as $\{(a,g)\} = F_D(d)$.

\noindent \textbf{Conquer with Task-Specific Agents}
After aspect-based thought grouping with Divider, three specialized agents collaborate to finalize the remaining ABSIA task for each thought group with batch prompt paradigm \cite{BatchPrompt-Lin-24ICLR}. 
The workflow branches into two independent streams: the Category Assignment Agent and the Opinion Term Extraction Agent. Subsequently, the Sentiment and Intensity Analysis Agent evaluates sentiment polarity and determines intensity scores for each thought group with opinion terms.
We describe the three specialized agents as follows:
\begin{itemize}[noitemsep, topsep=0pt, partopsep=0pt, parsep=0pt, leftmargin=*]
    \item \textbf{Category Assignment Agent}, denoted by $F_C$, assigns the most appropriate category $c$ to an aspect-based thought group $(a,g)$. This agent is equipped with a rule-based retrieval-augmented generation (RAG) \cite{li2025multilingual, adamu2025querying, triantafyllopoulos2025simple} tool that can return the predefined category list $\mathrm{RAG}(dm)$ \cite{MvpABSA-Gou-ACL23} based on domain information $dm$. We formalize the process as $c = F_C(a, g, \mathrm{RAG}(dm))$;

    \item \textbf{Opinion Extraction Agent}, denoted by $F_O$, extracts opinion words  or phrases that convey sentiment intensity for a thought group $(a,g)$. The opinion terms $\{o\}$ must be explicit substrings of the original text, preserving informal stylistic features such as lengthening expressions and punctuation). This is formalized as $\{o\} = F_O(a, g)$;

    \item \textbf{Sentiment Polarity \& Intensity Analysis Agent}, denoted by $F_S$, operates sequentially after the Opinion  Extraction Agent. It determines the sentiment polarity $s$ (positive/negative) for each aspect-opinion tuple with its corresponding thought group $(a, g, o)$. Additionally, it assigns a sentiment intensity score $i$ \cite{Service-Mamta-23ECIR} on a scale from 0 to 5 to indicate the strength of the expressed sentiment. The output is formalized as $(s, i) = F_S(a, g, o)$. 
\end{itemize}

\noindent \textbf{Merge and Structure Output}
We develop a rule-based integration method to merge outputs from the four specialized agents, producing the final output of a Dance team, denoted by $T$. This output is a structured list of aspect-based thought groups paired with corresponding ACOSI tuples, formally defined as $T = \{(g, a, c, o, s, i)\}$. We summarize the step-by-step Dance procedure in Algorithm \ref{alg:Dance}, and the overall process is formalized as
$\textit{T} = \mathrm{Dance} (d, dm)$.

\begin{algorithm}[t]
\caption{Dance}\label{alg:Dance}
\hspace*{\algorithmicindent} \textbf{Input:} Four LLM-based Agents $F_D$, $F_C$, $F_O$, $F_S$, a user review $d$ from domain $dm$, two rule-based tools $\mathrm{RAG}$, $\mathrm{Merge}$ \\
\hspace*{\algorithmicindent} \textbf{Output:} A list of aspect-based thought groups with ACOSI tuples $T$
\begin{algorithmic}[1]
\State Initialize $T \gets \{\}$
\State $\{(a, g)\} \gets F_D(d)$ \Comment{Divider}
\For{each $(a_j, g_j)$ in $\{(a, g)\}$}
    \State $c \gets F_C(a_j, g_j, \mathrm{RAG}(dm))$ \Comment{Category}
    \State $\{o\} \gets F_O(a_j, g_j)$ \Comment{Opinions extraction}
    \For{each $o_k$ in $\{o\}$}
        \State $(s, i) \gets F_S(a_j, g_j, o_k)$ \Comment{Sentiment}
        \State $t \gets \mathrm{Merge}((a_j, g_j), c, o_k, (s, i))$
        \State Add $t$ to $T$
    \EndFor
\EndFor
\State \textbf{Return} $T$
\end{algorithmic}
\end{algorithm}

\subsection{HA: Human-AI Collaboration for Annotation}
\noindent \textbf{Manager Agent for Automatic Annotation}
We employ an Manager Agent (MA) for preliminary automatic annotate $label$, which operates on the same output structure as $T$.
By varying base models and parameters for the LLM agents, we instantiate multiple Dance teams, 
each potentially producing diverse outputs for the same input document $d$, as illustrated in Algorithm \ref{alg:Dance}.  MA receives the task guidelines and the candidate outputs from all Dance teams, then integrates them by resolving inconsistencies and reconciling conflicts.
This results in a consensus-driven annotation grounded in the collective insights of multiple collaborative agents. 

\noindent \textbf{Human Revision} 
We recruited human annotators to evaluate and revise the automatically generated labels produced by two manager agents powered by the DeepSeek-V3 and GPT-4o backbone models, following protocols outlined in \cite{Lan-25ACL-UniT, Zhang-25NAACL-DynClean}.
Annotators received the original text and its candidate labels and were instructed to 1) keep high-quality labels, 2) correct inconsistencies, 3) discard low-confidence or conflicting labels, and 4) add new labels when the agents’ outputs were incomplete. In total, they reviewed 2,714 documents containing 45,304 ACOSI tuples: 13,716 unique tuples (21,152 instances) were retained, 8,233 revised, 1,075 newly added, and 16,994 discarded. The authors then performed a final proofread of the dataset. To this end, we curate the Inf-ABSIA dataset, comprising 2,714 documents and 23,024 ACOSI tuples annotated through Human-AI collaboration as gold labels.

\subsection{Knowledge Distillation with Reasoning Chain}
Given an input document $d$ and label $T = \{(g, a, c, o, s, i)\}$, we can construct a super-prompt $sp$ for reasoning by merging the prompts of all expert agents. The construction of reasoning chain $rc$ \cite{Zhang-ChainOpt-24NeurIPS, you2025probabilistic} perform the following step: 1) Aspect-based thought grouping. Output as $\{(a, g)\}$ 2) Category Assignment. Output as $\{(a, c)\}$ 3) Opinion extraction. Output as $\{(a, o)\}$ 4) Sentiment analysis. Output as $\{(a, o, s, i)\}$ 5) Merge results. Output as $\{(g, a, c, o, s, i)\}$. We can construct a reasoning dataset in the form $\{(d, sp, rc(T)), (d,T) \in \textbf{Inf-ABSIA}\}$.
We then distill knowledge \cite{jones2025dataset} of our proposed framework, DanceHA, into student models through supervised fine-tuning with LoRA \cite{Edward-LoRA-21arXiv}.

\begin{table}[t]
    \centering
    \resizebox{0.47\textwidth}{!}{
    \begin{tabular}{lllllccc}
        \toprule
        \textbf{Domain}  &\textbf{Method} & \textbf{Model} & \textbf{F1 (\%)} & \textbf{Acc (\%)} & \textbf{MAE} \\
        \midrule
        \multirow{11}{*}{Rest}  & zs-CoT & GPT-4o & 22.95 \footnotesize{$\pm$ 1.14} & 12.97 \footnotesize{$\pm$ 0.73} & 0.467 \footnotesize{$\pm$ 0.053} \\ 
                             & fs-CoT & GPT-4o & \textbf{35.04} \footnotesize{$\pm$ 2.43} & \textbf{21.27} \footnotesize{$\pm$ 1.80} & \textbf{0.428} \footnotesize{$\pm$ 0.023} \\ 
                             \cmidrule{2-6}
                             & Dance & GPT-3.5-turbo & 4.89 \footnotesize{$\pm$ 0.31} & 2.50 \footnotesize{$\pm$ 0.16} & 0.791 \footnotesize{$\pm$ 0.073} \\
                             & Dance & GPT-4o-mini & 19.90 \footnotesize{$\pm$ 0.38} & 11.05 \footnotesize{$\pm$ 0.23} & 0.418 \footnotesize{$\pm$ 0.007} \\
                             & Dance & GPT-4o & \textbf{47.80} \footnotesize{$\pm$ 1.65} & \textbf{31.42} \footnotesize{$\pm$ 1.41} & \textbf{0.210} \footnotesize{$\pm$ 0.019} \\
                             \cmidrule{2-6}
                             & Dance & DeepSeek-V3 & 37.93 \footnotesize{$\pm$ 1.75} & 23.42 \footnotesize{$\pm$ 1.33} & 0.274 \footnotesize{$\pm$ 0.016} \\
                             & Dance & Qwen3-4B & 13.68 \footnotesize{$\pm$ 1.13} & 7.34 \footnotesize{$\pm$ 0.65} & 0.909 \footnotesize{$\pm$ 0.051} \\
                             & Dance & Qwen3-14B & 25.87 \footnotesize{$\pm$ 1.85} & 14.87 \footnotesize{$\pm$ 1.23} & 1.016 \footnotesize{$\pm$ 0.010} \\
                             & Dance & Qwen2.5-72B & 35.38 \footnotesize{$\pm$ 1.23} & 21.50 \footnotesize{$\pm$ 0.91} & 0.364 \footnotesize{$\pm$ 0.007} \\
                             \cmidrule{2-6}
                             & MA & GPT-4o & 60.89 \footnotesize{$\pm$ 1.43} & 43.79 \footnotesize{$\pm$ 1.47} & 0.097 \footnotesize{$\pm$ 0.011} \\
                             & MA & DeepSeek-V3 & \textbf{63.18} \footnotesize{$\pm$ 1.62} & \textbf{46.20} \footnotesize{$\pm$ 1.72} & \textbf{0.086} \footnotesize{$\pm$ 0.010} \\
                                    
        \midrule
        \multirow{11}{*}{Hotel}  & zs-CoT & GPT-4o & 21.74 \footnotesize{$\pm$ 0.87} & 12.20 \footnotesize{$\pm$ 0.54} & 0.511 \footnotesize{$\pm$ 0.025} \\
                             & fs-CoT & GPT-4o & \textbf{31.40} \footnotesize{$\pm$ 2.03} & \textbf{18.64} \footnotesize{$\pm$ 1.42} & \textbf{0.453} \footnotesize{$\pm$ 0.007} \\
                             \cmidrule{2-6}
                             & Dance & GPT-3.5-turbo & 2.81 \footnotesize{$\pm$ 0.33} & 1.42 \footnotesize{$\pm$ 0.17} & 0.710 \footnotesize{$\pm$ 0.170} \\
                             & Dance & GPT-4o-mini & 16.48 \footnotesize{$\pm$ 1.80} & 8.99 \footnotesize{$\pm$ 1.07} & 0.458 \footnotesize{$\pm$ 0.008} \\
                             & Dance & GPT-4o & \textbf{44.18} \footnotesize{$\pm$ 1.30} & \textbf{28.36} \footnotesize{$\pm$ 1.07} & \textbf{0.251} \footnotesize{$\pm$ 0.006} \\
                             \cmidrule{2-6}
                             & Dance & DeepSeek-V3 & 37.47 \footnotesize{$\pm$ 1.51} & 23.06 \footnotesize{$\pm$ 1.14} & 0.315 \footnotesize{$\pm$ 0.008} \\
                             & Dance & Qwen3-4B & 14.91 \footnotesize{$\pm$ 2.14} & 8.07 \footnotesize{$\pm$ 1.25} & 1.192 \footnotesize{$\pm$ 0.045} \\
                             & Dance & Qwen3-14B & 24.34 \footnotesize{$\pm$ 1.24} & 13.86 \footnotesize{$\pm$ 0.80} & 1.124 \footnotesize{$\pm$ 0.018} \\
                             & Dance & Qwen2.5-72B & 32.76 \footnotesize{$\pm$ 2.17} & 19.61 \footnotesize{$\pm$ 1.54} & 0.327 \footnotesize{$\pm$ 0.002} \\
                            \cmidrule{2-6}
                             & MA & GPT-4o & 56.71 \footnotesize{$\pm$ 0.82} & 39.58 \footnotesize{$\pm$ 0.80} & 0.105 \footnotesize{$\pm$ 0.005} \\
                             & MA & DeepSeek-V3 & \textbf{59.07} \footnotesize{$\pm$ 1.92} & \textbf{41.94} \footnotesize{$\pm$ 1.92} & \textbf{0.088} \footnotesize{$\pm$ 0.008} \\
                                
        \midrule
        \multirow{11}{*}{Lap}  & zs-CoT & GPT-4o & 12.80 \footnotesize{$\pm$ 0.66} & 6.84 \footnotesize{$\pm$ 0.38} & 0.447 \footnotesize{$\pm$ 0.044} \\
                             & fs-CoT & GPT-4o & \textbf{16.61} \footnotesize{$\pm$ 0.71} & \textbf{9.06} \footnotesize{$\pm$ 0.42} & \textbf{0.465} \footnotesize{$\pm$ 0.034} \\
                            \cmidrule{2-6}
                             & Dance & GPT-3.5-turbo & 1.32 \footnotesize{$\pm$ 0.39} & 0.66 \footnotesize{$\pm$ 0.20} & 0.734 \footnotesize{$\pm$ 0.106} \\
                             & Dance & GPT-4o-mini & 6.91 \footnotesize{$\pm$ 0.27} & 3.58 \footnotesize{$\pm$ 0.15} & 0.451 \footnotesize{$\pm$ 0.059} \\
                             & Dance & GPT-4o & \textbf{29.97} \footnotesize{$\pm$ 0.78} & \textbf{17.63} \footnotesize{$\pm$ 0.54} & \textbf{0.215} \footnotesize{$\pm$ 0.008} \\
                             \cmidrule{2-6}
                             & Dance & DeepSeek-V3 & 26.68 \footnotesize{$\pm$ 1.10} & 15.40 \footnotesize{$\pm$ 0.73} & 0.295 \footnotesize{$\pm$ 0.006} \\
                             & Dance & Qwen3-4B & 5.05 \footnotesize{$\pm$ 0.33} & 2.59 \footnotesize{$\pm$ 0.17} & 0.998 \footnotesize{$\pm$ 0.099} \\
                             & Dance & Qwen3-14B & 11.49 \footnotesize{$\pm$ 0.96} & 6.10 \footnotesize{$\pm$ 0.54} & 0.842 \footnotesize{$\pm$ 0.093} \\
                             & Dance & Qwen2.5-72B & 16.85 \footnotesize{$\pm$ 0.80} & 9.20 \footnotesize{$\pm$ 0.47} & 0.356 \footnotesize{$\pm$ 0.030} \\
                             \cmidrule{2-6}
                             & MA & GPT-4o & 45.21 \footnotesize{$\pm$ 0.75} & 29.21 \footnotesize{$\pm$ 0.62} & \textbf{0.103} \footnotesize{$\pm$ 0.008} \\
                             & MA & DeepSeek-V3 & \textbf{47.40} \footnotesize{$\pm$ 1.21} & \textbf{31.07} \footnotesize{$\pm$ 1.04} & 0.105 \footnotesize{$\pm$ 0.006} \\
        \bottomrule
    \end{tabular}
    }
    \caption{Performance comparison of Dance, MA, zero-shot (zs) and 5-shot (fs) CoT with various base models on Inf-ABSIA. F1 and Acc represent the evaluation metrics for the ACOS tuple extraction, MAE for SIS. $\pm$ indicates std score. We use boldface to indicate the best performance within each method.}
    \label{tab:cot_vs_agent_metrics}
\end{table}

\begin{table*}[t]
    \centering
    \resizebox{0.9\textwidth}{!}{
    \begin{tabular}{llcccccccc}
    \toprule
 \multicolumn{2}{l}{\textbf{Expert Agents}} &  \multicolumn{3}{c}{\textbf{Divider}($F_D$)} & \multicolumn{1}{c}{\textbf{Category}($F_C$)} & \multicolumn{1}{c}{\textbf{Opinion}($F_O$)}  & \multicolumn{3}{c}{\textbf{Sentiment Polarity \& Intensity}($F_S$)}\\
 \cmidrule(lr){1-2} \cmidrule(lr){3-5} \cmidrule(lr){6-6} \cmidrule(lr){7-7} \cmidrule(lr){8-10} 
 
        \multicolumn{2}{l}{\textbf{Sub-tasks}} & \textbf{Thought Group} & \textbf{Aspect} & \textbf{All} & \textbf{All} & \textbf{All} & \textbf{Polarity} & \textbf{SIS} & \textbf{All} \\
        \midrule
        \multirow{7}{*}{Rest} 
                            & GPT-3.5-turbo & 17.71 \footnotesize{$\pm$ 0.16} & 8.54 \footnotesize{$\pm$ 1.22} & 2.40 \footnotesize{$\pm$ 0.26} & 75.33 \footnotesize{$\pm$ 1.88} & 36.87 \footnotesize{$\pm$ 3.42} & \textbf{96.69} \footnotesize{$\pm$ 0.80} & 40.06 \footnotesize{$\pm$ 2.76} & 38.73 \footnotesize{$\pm$ 2.68}  \\
                            & GPT-4o-mini & 20.32 \footnotesize{$\pm$ 0.19} & 29.66 \footnotesize{$\pm$ 1.48} & 10.29 \footnotesize{$\pm$ 0.28} & 78.37 \footnotesize{$\pm$ 1.29} & 46.77 \footnotesize{$\pm$ 2.86} & 95.20 \footnotesize{$\pm$ 1.98} & 59.52 \footnotesize{$\pm$ 0.48} & 56.66 \footnotesize{$\pm$ 1.22}   \\
                            & GPT-4o & \textbf{54.11} \footnotesize{$\pm$ 1.49} & \textbf{54.53} \footnotesize{$\pm$ 1.73} & \textbf{36.01} \footnotesize{$\pm$ 1.21} & \textbf{82.58} \footnotesize{$\pm$ 0.69} & \textbf{68.39} \footnotesize{$\pm$ 1.07} & 95.96 \footnotesize{$\pm$ 0.49} & \textbf{73.43} \footnotesize{$\pm$ 1.25} & \textbf{70.47} \footnotesize{$\pm$ 1.54}    \\
                            & DeepSeek-V3 & \underline{51.81} \footnotesize{$\pm$ 1.09} & 46.47 \footnotesize{$\pm$ 0.82} & \underline{28.34} \footnotesize{$\pm$ 0.35} & \underline{81.79} \footnotesize{$\pm$ 0.71} & \underline{61.06} \footnotesize{$\pm$ 0.93} & 95.14 \footnotesize{$\pm$ 1.64} & \underline{70.52} \footnotesize{$\pm$ 2.18} & \underline{67.09} \footnotesize{$\pm$ 2.43}  \\
                            & Qwen3-4B & 24.40 \footnotesize{$\pm$ 0.17} & 26.90 \footnotesize{$\pm$ 2.65} & 9.53 \footnotesize{$\pm$ 0.76} & 72.77 \footnotesize{$\pm$ 0.33} & 35.43 \footnotesize{$\pm$ 2.34} & \underline{96.59} \footnotesize{$\pm$ 1.97} & 39.01 \footnotesize{$\pm$ 2.88} & 37.69 \footnotesize{$\pm$ 3.10}  \\
                            & Qwen3-14B & 42.19 \footnotesize{$\pm$ 0.75} & 46.36 \footnotesize{$\pm$ 2.09} & 21.79 \footnotesize{$\pm$ 1.31} & 68.95 \footnotesize{$\pm$ 0.24} & 43.70 \footnotesize{$\pm$ 1.33} & 95.13 \footnotesize{$\pm$ 1.17} & 28.56 \footnotesize{$\pm$ 1.19} & 27.18 \footnotesize{$\pm$ 1.38}  \\
                            & Qwen2.5-72B  & 42.22 \footnotesize{$\pm$ 0.43} & \underline{48.73 }\footnotesize{$\pm$ 0.40} & 24.59 \footnotesize{$\pm$ 0.60} & 79.20 \footnotesize{$\pm$ 0.96} & 55.05 \footnotesize{$\pm$ 2.22} & 94.19 \footnotesize{$\pm$ 1.04} & 61.05 \footnotesize{$\pm$ 1.14} & 57.50 \footnotesize{$\pm$ 1.17}  \\

        \midrule
        \multirow{7}{*}{Hotel} 
                                & GPT-3.5-turbo & 17.26 \footnotesize{$\pm$ 0.65} & 5.07 \footnotesize{$\pm$ 0.50} & 1.96 \footnotesize{$\pm$ 0.08} & 75.96 \footnotesize{$\pm$ 2.30} & 39.30 \footnotesize{$\pm$ 1.54} & 93.42 \footnotesize{$\pm$ 1.80} & 41.29 \footnotesize{$\pm$ 8.63} & 38.53 \footnotesize{$\pm$ 7.99}   \\ 
                                & GPT-4o-mini & 21.40 \footnotesize{$\pm$ 1.18} & 24.31 \footnotesize{$\pm$ 1.83} & 9.46 \footnotesize{$\pm$ 0.63} & 72.29 \footnotesize{$\pm$ 0.99} & 50.37 \footnotesize{$\pm$ 2.47} & 93.72 \footnotesize{$\pm$ 0.60} & 57.47 \footnotesize{$\pm$ 1.19} & 53.85 \footnotesize{$\pm$ 0.76}   \\
                                & GPT-4o & \textbf{54.17} \footnotesize{$\pm$ 0.80} & \textbf{51.91} \footnotesize{$\pm$ 1.57} & \textbf{33.28} \footnotesize{$\pm$ 1.29} & \underline{77.68} \footnotesize{$\pm$ 0.84} & \textbf{68.19} \footnotesize{$\pm$ 2.13} & \underline{95.44} \footnotesize{$\pm$ 0.54} & \textbf{69.69} \footnotesize{$\pm$ 0.68} & \textbf{65.55} \footnotesize{$\pm$ 0.53} \\
                                & DeepSeek-V3 & \underline{52.01} \footnotesize{$\pm$ 1.18} & \underline{48.28} \footnotesize{$\pm$ 0.84} & \underline{29.31} \footnotesize{$\pm$ 1.02} & \textbf{78.27} \footnotesize{$\pm$ 1.72} & \underline{62.65} \footnotesize{$\pm$ 0.49} & 95.02 \footnotesize{$\pm$ 0.98} & \underline{65.34} \footnotesize{$\pm$ 0.79} & \underline{62.08} \footnotesize{$\pm$ 0.52}  \\
                                & Qwen3-4B & 29.02 \footnotesize{$\pm$ 0.53} & 29.52 \footnotesize{$\pm$ 2.49} & 12.34 \footnotesize{$\pm$ 0.85} & 66.25 \footnotesize{$\pm$ 0.50} & 39.91 \footnotesize{$\pm$ 2.76} & 94.20 \footnotesize{$\pm$ 0.86} & 27.88 \footnotesize{$\pm$ 1.10} & 26.27 \footnotesize{$\pm$ 1.25}  \\
                                & Qwen3-14B & 41.38 \footnotesize{$\pm$ 1.03} & 41.86 \footnotesize{$\pm$ 1.81} & 19.68 \footnotesize{$\pm$ 1.39} & 68.89 \footnotesize{$\pm$ 0.84} & 47.51 \footnotesize{$\pm$ 1.74} & \textbf{95.76} \footnotesize{$\pm$ 0.54} & 22.59 \footnotesize{$\pm$ 1.99} & 21.63 \footnotesize{$\pm$ 1.88}  \\
                                & Qwen2.5-72B & 40.98 \footnotesize{$\pm$ 0.83} & 47.38 \footnotesize{$\pm$ 1.20} & 23.94 \footnotesize{$\pm$ 0.70} & 74.26 \footnotesize{$\pm$ 1.10} & 56.73 \footnotesize{$\pm$ 0.54} & 94.54 \footnotesize{$\pm$ 1.04} & 64.08 \footnotesize{$\pm$ 1.02} & 60.58 \footnotesize{$\pm$ 1.04}  \\
                                             
        \midrule
        
        \multirow{7}{*}{Lap} 
                            & GPT-3.5-turbo & 14.08 \footnotesize{$\pm$ 1.34} & 7.16 \footnotesize{$\pm$ 0.60} & 1.84 \footnotesize{$\pm$ 0.42} & 32.04 \footnotesize{$\pm$ 3.40} & 34.12 \footnotesize{$\pm$ 2.72} & 94.27 \footnotesize{$\pm$ 4.10} & 37.18 \footnotesize{$\pm$ 4.47} & 35.21 \footnotesize{$\pm$ 5.74}  \\ 
                            & GPT-4o-mini & 15.01 \footnotesize{$\pm$ 1.97} & 19.12 \footnotesize{$\pm$ 1.01} & 5.06 \footnotesize{$\pm$ 0.57} & 40.76 \footnotesize{$\pm$ 0.54} & 46.67 \footnotesize{$\pm$ 0.69} & 93.72 \footnotesize{$\pm$ 0.84} & 54.52 \footnotesize{$\pm$ 4.58} & 51.08 \footnotesize{$\pm$ 4.10}  \\
                            & GPT-4o & \textbf{51.03} \footnotesize{$\pm$ 2.01} & \textbf{43.98} \footnotesize{$\pm$ 0.41} & \textbf{27.22} \footnotesize{$\pm$ 1.11} & \underline{64.02} \footnotesize{$\pm$ 0.48} & \textbf{61.29} \footnotesize{$\pm$ 0.85} & \textbf{94.86} \footnotesize{$\pm$ 0.49} & \textbf{73.79} \footnotesize{$\pm$ 0.31} & \textbf{70.00} \footnotesize{$\pm$ 0.33}  \\
                            & DeepSeek-V3 & \underline{50.68} \footnotesize{$\pm$ 1.31} & \underline{42.23} \footnotesize{$\pm$ 1.72} & \underline{24.51} \footnotesize{$\pm$ 0.64} & \textbf{65.79} \footnotesize{$\pm$ 2.74} & \underline{55.98} \footnotesize{$\pm$ 2.61} & 93.47 \footnotesize{$\pm$ 0.46} & \underline{68.68} \footnotesize{$\pm$ 1.44} & \underline{64.20} \footnotesize{$\pm$ 1.66}  \\
                            & Qwen3-4B & 22.81 \footnotesize{$\pm$ 0.98} & 23.19 \footnotesize{$\pm$ 0.88} & 7.09 \footnotesize{$\pm$ 0.37} & 34.50 \footnotesize{$\pm$ 0.91} & 31.07 \footnotesize{$\pm$ 1.38} & 93.09 \footnotesize{$\pm$ 1.48} & 38.13 \footnotesize{$\pm$ 4.08} & 35.54 \footnotesize{$\pm$ 4.23}  \\
                            & Qwen3-14B & 40.13 \footnotesize{$\pm$ 1.70} & 34.82 \footnotesize{$\pm$ 1.70} & 14.26 \footnotesize{$\pm$ 0.19} & 44.71 \footnotesize{$\pm$ 2.11} & 37.31 \footnotesize{$\pm$ 1.44} & \underline{94.72} \footnotesize{$\pm$ 1.34} & 35.69 \footnotesize{$\pm$ 5.45} & 33.88 \footnotesize{$\pm$ 5.65}  \\
                            & Qwen2.5-72B & 38.98 \footnotesize{$\pm$ 1.52} & 36.33 \footnotesize{$\pm$ 1.21} & 17.86 \footnotesize{$\pm$ 1.27} & 51.93 \footnotesize{$\pm$ 1.86} & 50.28 \footnotesize{$\pm$ 0.76} & 91.93 \footnotesize{$\pm$ 0.34} & 62.70 \footnotesize{$\pm$ 0.97} & 57.64 \footnotesize{$\pm$ 1.09}  \\
        \bottomrule
    \end{tabular}
    }
    \caption{Evaluation of Expert Agents in Dance and their sub-tasks. The evaluation metric is accuracy (\%). We use boldface to indicate the best performance and underline to indicate the second-best.
    }
    \label{tab:expert_agent_performance_all}
\end{table*}

\subsection{Evaluation Methods}
\label{sec:EvalMethod}
\noindent \textbf{Overall Performance}
Following prior studies \cite{MvpABSA-Gou-ACL23, CompoundABSALLM-Bai-24EMNLP, SelfConsistent-Kim-24ACL}, we adopt F1-score and Accuracy (Acc) as the primary evaluation metrics for the ACOS tuple extraction task. We also use Mean Absolute Error (MAE) for sentiment intensity score (SIS), which captures how well each baseline estimates SIS, especially in the presence of informal expressions. These 3 metrics (F1, Acc, MAE) together offer a comprehensive assessment of model performance on the ACOSI tuple extraction task. 

\noindent \textbf{Expert Agents and Sub-tasks Performance of Dance}
We use accuracy (Acc) as the evaluation metric for each expert agent and its associated sub-task. For convenience, we denote accuracy as $P$. Given the outputs produced by a Dance team and the ground-truth labels provided by Human-AI collaboration, we directly evaluate the performance of the Divider agent $F_D$ using $P(a, g)$.
 Additionally, we assess the sub-tasks of aspect extraction and thought grouping individually using $P(a)$ and $P(g)$, respectively.
To estimate the performance of downstream agents we use the probability chain rule \cite{bertsekas2008introduction}. Specifically, we assess the performance of $F_C$ as:
\begin{equation}
P(c \mid a, g) \approx P(c \mid a) = P(a, c)/P(a)
\label{eq:chain-rule}
\end{equation}
We make this approximation because a document can be segmented into thought groups in multiple ways. However, as long as the aspect extraction and the downstream predictions for opinion terms and sentiment are correct, the aspect-based thought grouping method can be considered valid. This is reasonable because ACOS evaluation metrics, such as F1-score, do not consider differences in thought group segmentation.
We apply the same approximation to $F_O$.
$F_S$ can be further decomposed into its two sub-tasks,
sentiment polarity and sentiment intensity analysis, which are evaluated as $P(s \mid a, g, o) \approx P(a, o, s)/P(a, o)$ and $P(i \mid a, g, o) \approx P(a, o, i)/P(a, o)$, respectively.

\section{Experiments}
\label{sec:exp_setup}
\subsection{Experimental Setting and Details}
Given the lack of comparable work on document-level ABSIA, we design several baselines to evaluate the performance of our DanceHA framework:
1) LLMs with zero-shot and few-shot (5-shot) Chain-of-Thought (CoT) Reasoning \cite{CoT-Jason-22CoRR}: We adapt ACOS tuple extraction prompts from prior studies \cite{CompoundABSALLM-Bai-24EMNLP, llama-Šmíd-24WASSA} and extend them with CoT prompting for document-level ACOSI tuple extraction. This baseline serves to isolate and assess the effectiveness of the Dance component. 
2) To examine the influence of the backbones to Dance, we implement the component with 7 LLMs: 3 close-source, GPT-4o, GPT-4o-mini, and GPT-3.5-turbo, and 4 open-source, DeepSeek-V3 \cite{Liu-DeepseekV3-24arXiv}, Qwen2.5-72B-Instruct \cite{Yang-Qwen25-24arXiv}, Qwen3-4B \cite{Qwen3-25arXiv} and Qwen3-14B.  
They allow us to analyze how different LLM backbones impact the overall ACOSI tuple extraction performance.
3) For MA, we use GPT-4o and DeepSeek-V3 as the backbone model due to their strong overall performance across tasks. We select 3 Dance teams as the inputs to MA with backbones as GPT-4o, DeepSeek-V3 and Qwen2.5-72B as their superior capability among the 7 LLMs used in this study.
4) We use Qwen3-4B and Qwen3-14B models for efficient reasoning fine-tuning with LoRA. The dataset is split into 70\% for training and 30\% for testing. We train for 3 epochs with a learning rate of 2e-4, a LoRA rank of 32, and a context length of 4096 tokens. We set the maximum token limit to 4096 for all LLMs and the temperature to 1.0 for all models, except for DeepSeek-V3, which is set to 0.6 following the official usage recommendations \cite{DeepSeek_R1_2025_arXiv}.
\subsection{Results}
\noindent \textbf{The effectiveness of Dance and MA.}
\label{sec:ValidityDance}
Table \ref{tab:cot_vs_agent_metrics} summarizes the performance of zero-shot and few-shot CoT method, Dance with the 7 LLMs, and MA with 2 advanced backbones. The main findings are as follows:
1) Dance equipped with GPT-4o reaches SOTA results among 7 backbones across 3 domains. This finding echoes prior work \cite{JudgingLLMs-Zheng-23NeurIPS} and reinforces the importance of choosing a strong base model for multi-agent frameworks.
2) Few-shot (5-shot) CoT with GPT-4o significantly outperforms zero-shot CoT, aligning with the findings of \cite{Zhang-SALLM-24NAACL}. However, it still lags behind our Dance equipped with GPT-4o and DeepSeek-V3.
3) Building on top of Dance teams and advanced LLMs, MA achieves a significant performance boost across 3 domains. For instance, in the restaurant domain, MA with DeepSeek-V3 reaches an F1 score of 63.18\%, substantially outperforming the best Dance with GPT-4o, which achieves 47.80\%. This finding highlights the importance of designing multi-agent collaboration topology for self-improvement \cite{Vighnesh-MultiagentFT-25ICLR}.

\begin{figure}[t]
\centering
\includegraphics[width=0.45\textwidth]{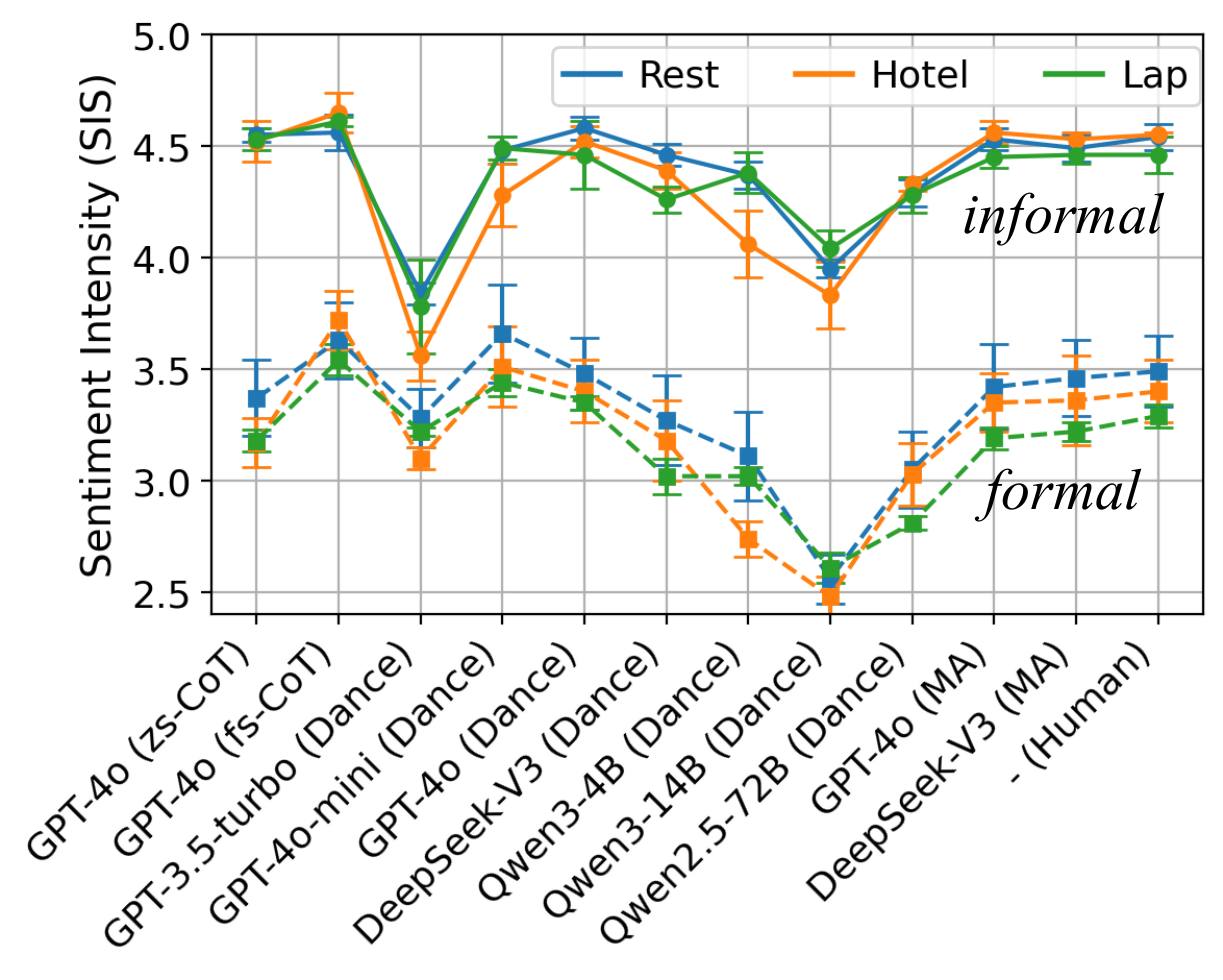}
\caption{SIS scores for ACOSI tuples with informal and formal styles among different baselines. 
Error bars indicate standard deviation score.
Results demonstrate the influence of informal styles on sentiment intensity and the capability of advanced models to interpret informal expressions. 
}
\label{fig:SIS}
\end{figure}

\noindent \textbf{Importance of Informal Styles.}
We analyze the differences in SIS between ACOSI tuples with and without informal styles (Figure~\ref{fig:SIS}). 
Across both CoT-based and agent-based methods, and consistently across all three domains, opinions expressed using informal styles (e.g., lengthening, extra punctuation) exhibit higher sentiment intensity score than those using more formal expressions. Notably, advanced models demonstrate a stronger sensitivity to opinions with informal cues, capturing more nuanced intensity variations in sentiment. 

\begin{table}[t]
    \Large
    \centering
    \resizebox{0.47\textwidth}{!}{
    \begin{tabular}{llccc}
        \toprule
 
         & \textbf{Method}  & \textbf{F1(\%)} & \textbf{Acc(\%)} & \textbf{MAE}  \\
        \midrule
        \multirow{3}{*}{Rest}  
                        & Dance  & 35.38   & 21.50  & 0.364  \\
                        
                        & w/o D\&C & 27.19  \textcolor{red}{$(8.19\downarrow)$}  & 15.73  \textcolor{red}{$(5.76\downarrow)$} & 0.473  \textcolor{red}{$(0.109\uparrow)$} \\
                        
                        & w/o Teamwork  & 32.48  \textcolor{red}{$(2.90\downarrow)$}  & 19.40 \textcolor{red}{$(2.10\downarrow)$} & 0.456  \textcolor{red}{$(0.093\uparrow)$} \\
                                   
        \midrule
        \multirow{3}{*}{Hotel} 
                        & Dance  & 32.76  & 19.61 & 0.327  \\
                        
                        & w/o D\&C  & 22.28 \textcolor{red}{$(10.48\downarrow)$}  & 12.54 \textcolor{red}{$(7.07\downarrow)$} & 0.490  \textcolor{red}{$(0.163\uparrow)$} \\
                        
                        & w/o Teamwork  & 29.81 \textcolor{red}{$(2.95\downarrow)$}  & 17.53 \textcolor{red}{$(2.08\downarrow)$} & 0.479  \textcolor{red}{$(0.151\uparrow)$} \\
        \midrule
        \multirow{3}{*}{Lap} 
                        & Dance  & 16.85  & 9.20 & 0.356  \\
                        
                        & w/o D\&C  & 10.42 \textcolor{red}{$(6.43\downarrow)$}  & 5.50 \textcolor{red}{$(3.70\downarrow)$} & 0.473  \textcolor{red}{$(0.118\uparrow)$} \\
                        
                        & w/o Teamwork  & 13.97 \textcolor{red}{$(2.88\downarrow)$}  & 7.51 \textcolor{red}{$(1.69\downarrow)$} & 0.476  \textcolor{red}{$(0.120\uparrow)$} \\
        \bottomrule
    \end{tabular}
    }
    \caption{Ablation results for the Dance component. w/o D\&C removes the Divide-and-Conquer Strategy; w/o Teamwork uses a single agent for the Conquer phase. 
    Red indicates performance drops.
    }
    \label{tab:ablation_study}
\end{table}

\noindent \textbf{Performance of Expert Agents.}
\label{sec:ExpertAgentsSubtask}
We present the accuracy of expert agents and their sub-tasks in Table \ref{tab:expert_agent_performance_all}.
1) The  Divider agent faces greater challenges, reflected by its relatively lower average accuracy, compared to other expert agents.
Moreover, the two sub-tasks handled by the Divider, aspect-based thought grouping and aspect terms extraction, are more difficult. This finding shows Divider agent has potential for further improvement.
2) While the sentiment polarity classification achieves consistently high accuracy  (average 94.62\%), SIS prediction remains challenging with an average Acc of 60.92\%. This indicates that LLMs struggle to interpret the nuanced sentiment intensity conveyed with informal expressions.
3) The Laptop domain has the lowest performance for Category Assignment agent due to its broad and diverse set of categories, which increases task complexity. Specifically, the number of categories is 121 in Laptop compared to 13 in Restaurant and 16 in Hotel.

\noindent \textbf{Distillation with Reasoning Chain.}
We present the reasoning distillation results with Qwen3-4B and Qwen3-14B in Table~\ref{tab:SFT-Qwen}. 
1) Efficient supervised fine-tuning on reasoning chains can effectively distill the knowledge of our DanceHA framework into student models. For instance, SFT Qwen-14B outperforms few-shot CoT GPT-4o across all domains and achieves performance comparable to Dance with Qwen2.5-72B. 
2) We configure Dance with Qwen3 models as non-thinking to explore whether multi-agent collaboration in Dance can substitute for inference with explicit reasoning. As noted in the Qwen3 technical report \cite{Qwen3-25arXiv}, enabling thinking mode equips the model with multi-step reasoning capabilities for complex tasks. Surprisingly, the results depend on model size. For the Qwen-4B, zero-shot CoT with reasoning outperforms Dance. However, for the larger Qwen-14B, Dance achieves comparable or better performance than zero-shot CoT with reasoning.

\begin{table}[t]
    \Large
    \centering
    \resizebox{0.47\textwidth}{!}{
    \begin{tabular}{llllllccc}
        \toprule
          &\textbf{Method} & \textbf{Model} & \textbf{Mode} & \textbf{F1 (\%)} & \textbf{Acc (\%)} & \textbf{MAE} \\
        \midrule
        \multirow{6}{*}{Rest}  
                             & zs-CoT & 4B & think & 20.22 \footnotesize{$\pm$ 2.34} & 11.27 \footnotesize{$\pm$ 1.44} & 1.238 \footnotesize{$\pm$ 0.068} \\
                             & zs-CoT & 14B & think & 19.06 \footnotesize{$\pm$ 2.80} & 10.56 \footnotesize{$\pm$ 1.73} & 0.740 \footnotesize{$\pm$ 0.100} \\
                             & Dance & 4B & none & 13.68 \footnotesize{$\pm$ 1.13} & 7.34 \footnotesize{$\pm$ 0.65} & 0.909 \footnotesize{$\pm$ 0.051} \\
                             & Dance & 14B & none & 25.87 \footnotesize{$\pm$ 1.85} & 14.87 \footnotesize{$\pm$ 1.23} & 1.016 \footnotesize{$\pm$ 0.010} \\
                             & SFT & 4B & think & \underline{30.82} \footnotesize{$\pm$ 1.14} & \underline{18.22} \footnotesize{$\pm$ 0.80} & \underline{0.408} \footnotesize{$\pm$ 0.026} \\
                             & SFT & 14B & think & \textbf{34.52} \footnotesize{$\pm$ 1.23} & \textbf{20.87} \footnotesize{$\pm$ 0.91} & \textbf{0.395} \footnotesize{$\pm$ 0.031} \\
                                                                        
        \midrule
        \multirow{6}{*}{Hotel}  
                             & zs-CoT & 4B & think & 20.38 \footnotesize{$\pm$ 1.27} & 11.35 \footnotesize{$\pm$ 0.79} & 1.165 \footnotesize{$\pm$ 0.176} \\
                             & zs-CoT & 14B & think & 24.56 \footnotesize{$\pm$ 1.67} & 14.01 \footnotesize{$\pm$ 1.09} & 0.940 \footnotesize{$\pm$ 0.058} \\
                             & Dance & 4B & none & 14.91 \footnotesize{$\pm$ 2.14} & 8.07 \footnotesize{$\pm$ 1.25} & 1.192 \footnotesize{$\pm$ 0.045} \\
                             & Dance & 14B & none & 24.34 \footnotesize{$\pm$ 1.24} & 13.86 \footnotesize{$\pm$ 0.80} & 1.124 \footnotesize{$\pm$ 0.018} \\
                             & SFT & 4B & think & \underline{30.38} \footnotesize{$\pm$ 3.53} & \underline{17.96} \footnotesize{$\pm$ 2.42} & \underline{0.462} \footnotesize{$\pm$ 0.038} \\
                             & SFT & 14B & think & \textbf{32.95} \footnotesize{$\pm$ 1.86} & \textbf{19.74} \footnotesize{$\pm$ 1.33} & \textbf{0.418} \footnotesize{$\pm$ 0.042} \\
                                        
        \midrule
        \multirow{6}{*}{Lap}  
                             & zs-CoT & 4B & think & 9.16 \footnotesize{$\pm$ 1.12} & 4.80 \footnotesize{$\pm$ 0.62} & 1.209 \footnotesize{$\pm$ 0.250} \\
                             & zs-CoT & 14B & think & 11.95 \footnotesize{$\pm$ 1.27} & 6.36 \footnotesize{$\pm$ 0.72} & 0.820 \footnotesize{$\pm$ 0.153} \\ 
                             & Dance & 4B & none & 5.05 \footnotesize{$\pm$ 0.33} & 2.59 \footnotesize{$\pm$ 0.17} & 0.998 \footnotesize{$\pm$ 0.099} \\
                             & Dance & 14B & none & 11.49 \footnotesize{$\pm$ 0.96} & 6.10 \footnotesize{$\pm$ 0.54} & 0.842 \footnotesize{$\pm$ 0.093} \\
                             & SFT & 4B & think & \underline{13.92} \footnotesize{$\pm$ 0.87} & \underline{7.48} \footnotesize{$\pm$ 0.50} & \underline{0.472} \footnotesize{$\pm$ 0.007} \\
                             & SFT & 14B & think & \textbf{17.19} \footnotesize{$\pm$ 0.79} & \textbf{9.40} \footnotesize{$\pm$ 0.47} & \textbf{0.452} \footnotesize{$\pm$ 0.050} \\         
        \bottomrule
    \end{tabular}
    }
    \caption{Results for distillation with reasoning chains. We use boldface to indicate the best performance and underline to indicate the second-best. Model 4B denotes Qwen3-4B; 14B denotes Qwen3-14B.}
    \label{tab:SFT-Qwen}
\end{table}

\noindent \textbf{Ablation Study}
\label{sec:AblationStudy}
We conduct ablation experiments to evaluate the effectiveness of key modules of Dance. Specifically, we design two experimental conditions: 1) Without Divide-and-Conquer Strategy (w/o D\&C)--we create a single agent baseline equipped with a super-prompt, formed by merging the prompts of all expert agents, excluding the aspect-based thought grouping step. Unlike the zero-shot CoT baseline in Table \ref{tab:cot_vs_agent_metrics}, this prompt includes a detailed definition of aspect term.
2) Without Teamwork in the Conquer Phase (w/o Teamwork): we combine prompts from the 3 expert agents in the Conquer phase into a single agent configuration, removing the collaborative setup.
GPT-4o is the underlying LLM for all experiments.
Table \ref{tab:ablation_study} summarizes the results.
We observe that the D\&C Strategy is critical to Dance's success, removing it results in a significance performance decline: with the average F1-score dropping by 8.37\%, accompanied by similar decreases in ACOS Accuracy and SIS prediction MAE.
As expected, the w/o Teamwork variant outperforms the w/o D\&C group, but still experiences a performance drop, with the average F1-score decreasing by 2.91\%.
This underscores the necessity of decomposing tasks in the Divider phase, enabling expert agents to handle manageable sub-tasks and mitigating information overload.

\noindent \textbf{Human Evaluation for Data Quality}
\label{sec:DataQuality}
We recruited two annotators to evaluate the quality of the labels generated by DanceHA.
Annotators were instructed to rate score on a 1–5 scale (1 = completely disagree, 5 = completely agree) 
for 90 randomly sampled documents (30 per domain) with a total of 820 ACOSI tuples. 
Inter-annotator agreement (IAA) is measured using Cohen’s Kappa Score \cite{cohen_kappa_score-Artstein-08ACL}.
We achieved an average rating score of 4.47 out of 5 and an IAA score of 0.61.
These results indicate DaceHA’s effectiveness in balancing fine-grained labeling with high annotation quality.

\begin{table}[t]
    \centering
    \resizebox{0.31\textwidth}{!}{
    \begin{tabular}{lcccc}
       \toprule
          & \textbf{Rest} & \textbf{Hotel} & \textbf{Laptop} & \textbf{Avg}\\
        \midrule
         Rater & 4.62 & 4.55 & 4.25 & 4.47 \\
         IAA & 0.65 & 0.61 & 0.58 & 0.61 \\
        \bottomrule
    \end{tabular}
    }
    \caption{Human evaluation results for data quality.}
    \label{tab:human_llm_eval}
\end{table}

\section{Conclusion}
This work presented DanceHA, a multi-agent framework for open-ended, document-level ABSIA with informal styles. 
Extensive experiments across three domains and seven LLMs demonstrate that Dance effectively handles complex ABSIA tasks and significantly outperforms few-shot CoT. 
Moreover, MA achieves even better performance by building upon Dance teams with different backbone models.
Our findings reveal the importance of informal styles in ABSIA, which often amplify sentiment intensity for specific aspects and opinions.
Our experiments of distillation with reasoning chain show that we can effectively transfer the multi-agents knowledge in DanceHA into student models. 

\section{Limitations}
We acknowledge several limitations for future exploration. First, the influence of team size on annotation quality remains an open question. Moreover, some sub-tasks might be replaced with small language models such as T5 and RoBERTa to improve efficiency.

\section{Acknowledgments}
This work was supported by the National Science Foundation awards III-2107213 and ITE-2333789.

\bibliography{bibs/ref_all}

\end{document}